\documentclass[12pt]{article}
\usepackage{amssymb,amsmath,cite,bm}
\usepackage[dvips]{graphicx,color}
\usepackage{psfrag,subfigure}
\newtheorem{theorem}{Theorem}
\newtheorem{remark}{Remark}
\newtheorem{proposition}{Proposition}

\newtheorem{assumption}{Assumption}

\begin{document}

% ----------------------------------------------------------------
\title{\bf Hybrid Simulator for Space Docking and Robotic Proximity Operations}

\author{Farhad Aghili\thanks{email: faghili@encs.concordia.ca}}

\date{}

\maketitle

%-------------------------------------------------------------
\begin{abstract}
In this work, we present a hybrid simulator for space docking and robotic proximity operations methodology. This methodology also allows for the emulation of a target robot operating in a complex environment by using an actual robot. The emulation scheme aims to replicate the dynamic behavior of the target robot interacting with the environment, without dealing with a complex calculation of the contact dynamics. This method forms a basis for the task verification of a flexible space robot. The actual emulating robot is structurally rigid, while the target robot
can represent any class of robots, e.g., flexible, redundant, or space robots. Although the emulating robot is not dynamically equivalent to the target robot, the dynamical similarity can be achieved by using a control law developed herein. The effect of disturbances and actuator dynamics on the fidelity and the contact stability of the robot emulation is thoroughly analyzed.
\end{abstract}
%-------------------------------------------------------------
%\cite{Aghili-Piedboeuf-2006} 
%\cite{Aghili-Piedboeuf-2003a}   
%\cite{Aghili-2005}                                  % TRO-2005
%\cite{Aghili-Su-2012}

%\cite{Aghili-2005b}                                 % IROS-2005
%\cite{Aghili-2012b}
%\cite{Aghili-2010}  
%\cite{Aghili-2010n}
%\cite{Aghili-Namvar-2008,Aghili-Namvar-Vukovich-2006}                           % IJRR-2009
%\cite{Aghili-Parsa-2009b}                           % TRO-2009
%\cite{Aghili-2009b}
%\cite{Aghili-2006}
%\cite{Aghili-Parsa-2008b}
%\cite{Aghili-2006b}                                 % ICRA-2006                               % ICRA-2006
%\cite{Aghili-2005b}
%\cite{Doyon-Piedboeuf-Aghili-Gonthier-Martin-2003}
%\cite{Piedboeuf-Aghili-Doyon-Martin-2002, Aghili-Piedboeuf-2002}   
%\cite{Piedboeuf-deCarufel-Aghili-Dupuis-1999}
%\cite{Aghili-2011k,Aghili-Parsa-2007b}      TRO paper
%\cite{Aghili-Kuryllo-Okouneva-English-2010a}
%\cite{Aghili-Parsa-2009}
%\cite{Aghili-2008c,Aghili-Parsa-Martin-2008a}
%\cite{Aghili-Salerno-2016}
%\cite{Aghili-2010s}
%\cite{Aghili-2016c}
%\cite{Aghili-2010p}
%\cite{Aghili-2010f}
%\cite{Aghili-Parsa-2008b}
%\cite{Aghili-Kuryllo-Okuneva-McTavish-2009}
%\cite{Aghili-Kuryllo-Okouneva-English-2010c}

%-------------------------------------------------------------
\section{Introduction and Motivation}
%-------------------------------------------------------------

There is  growing interest in employing robotic
manipulators to perform assembly tasks. In particular, space robots
have become a viable means to perform extra-vehicular robotic tasks
\cite{Aghili-Piedboeuf-2006,Skaar-Ruoff-Seebass-1994,Aghili-2010n,Aghili-Su-2012,Aghili-2012b,Piedboeuf-Aghili-Doyon-Martin-2002, Aghili-Piedboeuf-2002,Aghili-2011k,Aghili-Salerno-2016,Aghili-Parsa-2007b}. For instance, the Special Purpose
Dextrous Manipulator (SPDM) will be used to handle various orbital
replacement units (ORU) for  maintenance operations performed on the
International Space Station.

The complexity of a robotic operation  associated with a task
demands a verification facility on the ground to ensure that the
task can be performed on orbit as intended. This requirement has
motivated the development of the SPDM task verification facility at
Canadian Space Agency
\cite{Piedboeuf-Dupuis-deCarufel-1998,Aghili-Dupuis-Piedboeuf-deCarufel-1999,Aghili-2019c,
Piedboeuf-Aghili-Doyon-Gonthier-Martin-Zhu-2001,
Piedboeuf-deCarufel-Aghili-Dupuis-1999,deCarufel-Martin-Piedboeuf-2000,Aghili-Piedboeuf-2002}.
The verification operation on the ground is challenging as space
robots are designed to work only in a micro-g environment. Actually,
thanks to weightlessness on orbit, the SPDM can handle payloads as
massive as $600 \; kg$. On the ground, however, it cannot even
support itself against gravity. Although this problem could be fixed
by using a system of weights and pullies to counterbalance gravity,
the weights change the robot inertia and its dynamic behavior.
Moreover, it is difficult to replicate the dynamic effects induced
by the flexibility of the space robot or space structure on which
the manipulator is stowed.

Simulation is another tool that can be used to validate the
functionality of a space manipulator
\cite{Ma-Buhariwala-Roger-MacLean-Carr-1997}. Although the dynamics
and kinematics models of space manipulators are more complex that
those of terrestrial manipulators due to dynamic coupling between
the manipulator and its spacecraft, Vafa {\em et al.}
\cite{Vafa-Dubowsky-1990,Vafa-Dubowsky-1990b} showed that the
space-robot dynamics can be captured by the concept of a {\em
Virtual-Manipulator}, which has a fixed base in inertial space at a
point called  {\em Virtual Ground}. Today, with the advent of
powerful computers, faithful and real-time simulators exist, which
are able to capture dynamics of almost any type of manipulator with
many degrees of freedom. However, simulation of a robot interacting
with its environment with a high fidelity requires accurate
modelling of both the manipulator and the contact interface between
the manipulator and its environment. Faithful models of space-robots
are available; it is the calculation of the contact force which
poses many technical difficulties associated with contact dynamics
\cite{Ma-Nahon-1992}. In the literature, many models for the contact
force, comprised of normal and friction forces, have been reported
\cite{Lankarani-Nikravesh-1988,Ma-Nahon-1992,Johnson-1985}. Hertzian
contact theory is used to estimate local forces \cite{Johnson-1985}.
However, predicting the contact point locations requires calling an
optimization routine \cite{Ma-Nahon-1992}, thus demanding
substantial computational effort. In particular, the complexity of
contact-dynamics modelling tends to increase exponentially when the
two objects have multi-point contacts
\cite{Ma-Buhariwala-Roger-MacLean-Carr-1997}. Moreover, since the
SPDM is tele-operated by a human operator, the validation process
requires inclusion of a real-time simulation environment since the
simulation should allow human operators to drive the simulation in
real-time. All these difficulties can be avoided by emulating the
contact dynamics.

%=============================================================
\begin{figure}
\centering{\includegraphics[width=10cm]{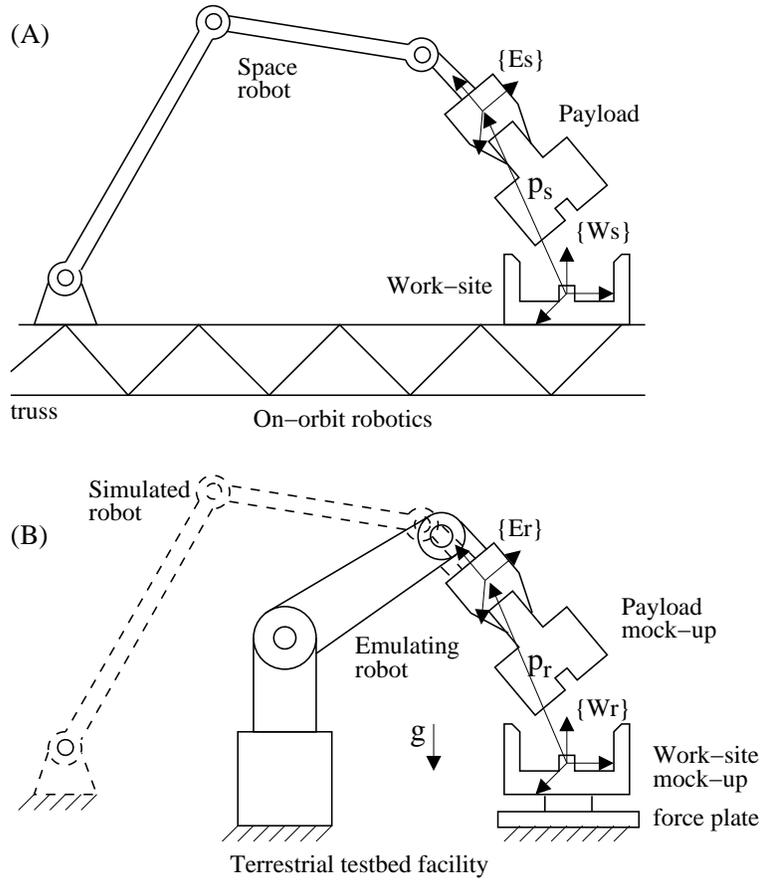}}
\caption{Schematics of the space manipulator (A) and the
terrestrial manipulator (B) performing a task involving contact.}
\label{fig_robots}
\end{figure}
%=============================================================

Design and implementation of a laboratory test-bed for study the
dynamics coupling between a space-manipulator and its spacecraft
operating in free space are presented in
\cite{Dubowsky-Durfee-Kuklinski-1994}. Similar concepts have been
also pursued by other space agencies  for different  applications
\cite{Krenn-Schafer-1999,Ananthakrishnan-Teders-Alder-1996,Aghili-2006b,Tarao-Inohira-Uchiyama-2000,Aghili-Namvar-2008,Aghili-2005b,Aghili-Namvar-Vukovich-2006,Aghili-Parsa-2009,Aghili-2006,Aghili-Parsa-Martin-2008a,Aghili-2016c,Aghili-2008c,Aghili-2016c,Aghili-2010f}.
In this work we develop an emulation
technique which can replicate the dynamic behavior of a robot
interacting with a complex environment, without actually simulating
the environment \cite{Aghili-Piedboeuf-2006}. Fig.\ref{fig_robots}B schematically illustrates the
concept of the robot emulation system. The system  consists of three
main functional elements: a real-time simulator capturing the
dynamics of the space robot, the mock-up of the payload and its
work-site, and a rigid robot which handles the payload.
Interconnecting the simulator and the emulating robot, the entire
system works in a {\em hardware-in-loop simulation} configuration
that permits us to replicate the dynamic behavior of the space robot
\cite{Aghili-Piedboeuf-2006,Aghili-Dupuis-Piedboeuf-deCarufel-1999,Aghili-Piedboeuf-2002}.

The purpose of the control system is to make the constrained-motion
dynamical behavior of the emulating robot as close as possible to
that actually encountered by the space robot \cite{Aghili-Piedboeuf-2006}. The obvious advantage
of this methodology is that there is no need for modeling a complex
environment. Moreover, because of using the real hardware, a visual
cue of the payload and the work-site is available which otherwise,
must be simulated. The main contribution of this paper is to layout
a control architecture for the emulation of a robot operating in
free space and in contact with the environment. It follows by an
analysis of the fidelity and the stability of the robot emulation;
the validity of the analysis is demonstrated by experimental
results.

This paper is organized as follows: In section \ref{sec_control}, we
formulate the emulation problem and develop the control law. Section
\ref{sec_flexible} describes the equations of motion of independent
coordinate which arise in the case of flexible target robots. A
thorough analysis of the robot emulation, including fidelity of the
emulation with respect to the disturbance and stability conditions
are given in section \ref{sec_Analysis}. 

%-------------------------------------------------------------
\section{Control of Emulating Robot}
\label{sec_control}
%-------------------------------------------------------------

Let's consider an emulating robot and a target robot, referred to
by subscripts $r$ and $s$, with $n$ and $m$ degrees-of-freedom
(DOF), respectively, operating in an $n$-dimensional task space.
The generalized coordinates of the robots are represented by
vectors $q_r \in \mathbb{R}^n$ and $q_s \in \mathbb{R}^m$,
respectively. Note that the emulating robot is rigid but the space
robot is typically a flexible one with many degrees-of-freedom.
Hence, we can say
\[ \mbox{dim}(q_s)=m \geq \mbox{dim}(q_r)=n. \]

Let vector $p_r^T=[x_r^T , \phi_r^T]$ denote the {\em pose} of the
end-effector of the emulating robot, where $x_r$ and $\phi_r$
describe, respectively, the position and a minimal representation of
the orientation of the end-effector expressed in the base frame
attached to the work-site; see Fig.\ref{fig_robots}. Similarly, the
pose of the target robot is denoted by vector $p_s$.

\begin{assumption} \label{ass_pose}

\begin{enumerate}
\item  Both robots have identical end-effector pose configuration
with respect to their base frames, i.e., $p_r = p_s = p$ and

\item they experience the same contact force/moment $f \in
\mathbb{R}^n$.
\end{enumerate}
\end{assumption}

Moreover, we assume that $M_r \in \mathbb{R}^{n \times n}$ and
$M_s \in \mathbb{R}^{m \times m}$ are the inertia matrices of the
two manipulators; $h_r \in \mathbb{R}^n$ and $h_r \in
\mathbb{R}^m$ contain Coriolis, centrifugal, friction, gravity
(for the emulating robot), and stiffness (for the space robot);
$\tau_r \in \mathbb{R}^n$ and $\tau_s \in \mathbb{R}^{n}$  are the
joint torque vectors; and that  $J_r =\partial p_r / \partial q_r
\in \mathbb{R}^{ n \times n}$ and $J_s = \partial p_s / \partial
q_s \in \mathbb{R}^{ n \times m}$ are the {\em analytical}
Jacobian matrices. Then, the equations of motion of the robots can
be described by
\cite{Canudas-Siciliano-Bastin-book-1996,Craig-1989,deJalon-Bayo-1989}
\begin{equation} \label{eq_GroundRobot}
M_r \ddot q_r + h_r(q_r, \dot q_r) = \tau_r + J_r^T f_a,
\end{equation}
\begin{equation} \label{eq_SpaceRobot}
M_s \ddot q_s + h_s(q_s, \dot q_s) = B \tau_s + J_s^T f_a,
\end{equation}
where $B= \left[\begin{array}{c} I \\ 0 \end{array} \right] \in
\mathbb{R}^{m \times n}$, and $f_a \in \mathbb{R}^n$ is the vector
of generalized constraint force expressed in the base frame. Note
that $f_a$ performs work on $\dot p$, and it is related to the
contact force $f$ by
\begin{equation} \label{eq_fa}
f_a= \Lambda(\phi) f,
\end{equation}
where $\Lambda (\phi)= \mbox{diag}\{ I , \Lambda_1(\phi) \}$ is a
transformation matrix that depends on a particular set of parameters
used to represent the end-effector orientation
\cite{Canudas-Siciliano-Bastin-book-1996}; $\phi_s=\phi_r=\phi$
because of Assumption \ref{ass_pose}.

Now, assume that a real-time simulator captures the dynamics of the
space robot. Also, assume that the endpoint pose of the simulated
robot is constrained by that of the emulating robot, i.e., the
following {\em rheonomic} constraint
\begin{equation} \label{eq_constraint}
{\Phi}({q}_s, {q}_r(t)) = p_r(q_r(t)) - p_s(q_s) \equiv 0
\end{equation}
holds. Enforcing the constraint ensures that the end-effector of the
simulated robot is constrained by the pose of the environment
barriers, which physically constrains the emulating robot  \cite{Aghili-2005,Aghili-Piedboeuf-2003a,Aghili-Parsa-2008b}. The
Lagrange equations of the simulated robot can be formally expressed
by a set of Differential Algebraic Equations (DAE) as
\begin{equation} \label{eq_SimRobot}
M_s \ddot{q}_s  + {\Phi}_{q_s}^T \lambda = - h_s + B\tau_s.
\end{equation}
\begin{equation} \label{eq_Phi}
{\Phi}(q_s, t) = 0
\end{equation}
where ${\Phi}_{qs}= \partial {\Phi} / \partial q_s =-J_s \in
\mathbb{R}^{m\times n}$ denotes the Jacobian of the constraint
with respect to $q_s$, and $\lambda \in \mathbb{R}^n$ represents
the Lagrange multiplier which can be solved from equation
(\ref{eq_SimRobot}) together with the second time derivative of
the constraint equation (\ref{eq_Phi}). That is,
\begin{equation} \label{eq_Lambda}
{\lambda} = (J_s M_s^{-1}{J}_s^T )^{-1} \left( c + J_s M_s^{-1}
(h_s - B \tau_s)  \right),
\end{equation}
where $c=\ddot p_r(t) - \dot J_s \dot q_s$; note that the
constraint force $\lambda$ can be measured only if  a measurement
of pose acceleration $\ddot p_r$ is available, say via an
accelerometer.

A comparison of equations (\ref{eq_SimRobot}) and
(\ref{eq_SpaceRobot}) reveals that the Lagrange multiplier
${\lambda}$ acts as a constraint force/moment on the simulated
robot. Therefore, our goal is to equate the constraint force and
the Lagrange multiplier, i.e., $f_a = {\lambda}$. This can be
achieved by using the {\em computed torque} method in the task
space
\cite{Craig-1989,Spong-Vidasagar-1989,Canudas-Siciliano-Bastin-book-1996,Aghili-Parsa-2009,Aghili-2010}.
Substituting the joint acceleration from (\ref{eq_GroundRobot})
into the second time derivative of the pose, $\ddot p_r = J_r
\ddot q_r + \dot J_r \dot q_r$, yields
\begin{equation}\label{eq_ddpose}
\ddot p_r= J_r M_r^{-1}(\tau_r + J_r^T f_a - h_r) + \dot J_r \dot
q_r.
\end{equation}
This equation can then be substituted into (\ref{eq_Lambda}) to
obtain the joint torque solution, assuming $f_a = \lambda$. That
is
\begin{eqnarray} \nonumber
\tau_r^c &=&  h_r + M_r J_r^{-1} ( J_s M_s^{-1} (B \tau_s -
h_s) + \dot J_s \dot{q}_s - \dot J_r \dot q_r ) \\
\label{eq_tau} & & - \underbrace{J_r^T ( I - {\cal M}_r {\cal
M}_s^{-1} ) f_a}_{\mbox{force feedback}},
\end{eqnarray}
where $ {\cal M} := \left[ J M^{-1} J^T \right]^{-1}$ is the {\em
Cartesian inertia matrix} whose inverse can be interpreted as
mechanical impedance. The torque control (\ref{eq_tau}) can be
grouped into force and motion feedback terms.
\begin{remark} \label{rm_equalmass}
The force feedback gain tends to decrease when the inertia ratio
approaches one. In the limit, i.e., when ${\cal M}_s= {\cal M}_r$,
the force feedback  is disabled, and hence the robot emulation can
be realized without using force feedback.
\end{remark}

In free space, where $f_a=0$, the controller matches the end-point
poses of two robots at acceleration level and that inevitably
leads to drift. We can improve the control law by incorporating a
feedback loop to minimize the drift error, i.e.,
\begin{eqnarray} \label{eq_tau'}
\tau_r^c &=&  h_r - M_r J_r^{-1} \dot{J}_r \dot{q}_r - J_r^T f_r +
M_r J_r^{-1} u
\end{eqnarray}
where
\begin{equation} \label{eq_u}
u :=  \tilde{a}_s  - G_v ( J_r \dot{q}_r - \int \tilde{a}_s dt )  -
G_p ( p_r(q_r) - \int \int \tilde{a}_s dt)
\end{equation}
is the auxiliary input, $\tilde{a}_s := J_s M_s^{-1} (B \tau_s- h_s)
+ \dot{J}_s \dot{q}_s + J_s M_s^{-1} J_s^T f_a$ is the estimated
pose acceleration of the target robot calculated from the
unconstrained dynamics model, and $G_v>0$ and $G_p>0 $ are drift
compensation gains. In the following, we show that the constraint
force asymptotically tracks the Lagrange multiplier. Denoting
\[ \ddot{e}_p := \ddot p_r - \tilde{a}_s, \]
then the dynamics of the errors under the control law
(\ref{eq_tau'}) is
\begin{equation} \label{eq_ddep}
\ddot{e}_p + G_v \dot{e}_p + G_p e_p = 0.
\end{equation}
The dynamics is asymptotically stable if the gain parameters are
adequately chosen, i.e., $e_p, \dot{e}_p , \ddot{e}_p \rightarrow 0$
as $t \rightarrow \infty$. Furthermore, let
\begin{equation} \label{eq_ef1}
e_f:=f_a - \lambda,
\end{equation}
represent the force error. Then, one can show from
(\ref{eq_Lambda}) that the force error is related to the
acceleration error by
\begin{equation} \label{eq_ef}
e_f = {\cal M}_s \ddot{e}_p.
\end{equation}
Since the mass matrix is bounded, i.e. $ \exists \kappa
>0 $ such that ${\cal M}_s(q_s) \leq \kappa I$, then
\[ \| e_f \| \leq \kappa \|\ddot {e}_p \| \]
and hence $e_f \rightarrow 0$ as $t \rightarrow \infty$.
Therefore, the error between the constraint force and the Lagrange
multiplier is asymptotically stable under the control law
(\ref{eq_tau'}). The above mathematical development can be
summarized in the following proposition.

\begin{proposition} \label{propos_constraint}
Assuming the following: (i) the endpoint pose of the simulated robot
is constrained with that of the emulating robot performing a contact
task, (ii) the torque control law (\ref{eq_tau'}) is applied on the
emulating robot. Then, (i) the emulating robot produces the
end-effector motion of the space robot, (ii) the constraint force
asymptotically approaches the Lagrange multiplier: $f_a \rightarrow
\lambda$.
\end{proposition}

In this case, the simulated robot behaves as if it interacts
virtually with the same environment as the emulating robot does.
In other words, the combined system of the simulator and the
emulating robot is dynamically equivalent to the original target
robot.

\subsection{Force Error Feedback}
Incorporating a feedback of the force error into the control law
(\ref{eq_tau'})-(\ref{eq_u}) can improve the force response of the
emulating system, particularly during the non-contact to contact
transition. However, to establish such a feedback requires an
estimation of the force error, which, in turn, can be obtained from
(\ref{eq_ef}) provided that a measurement of the pose acceleration
is available. Then one can show that changing the auxiliary input
(\ref{eq_u}) to $u=\tilde a_s - G_v \dot e_p -G_p e_p - G_f e_f$,
where $G_f>0$ is the force feedback gain, yields the following error
dynamics
\[ [I+ {\cal M}_s G_f ] \ddot e_p + G_v \dot e_p + G_p e_p =0.\]
The above differential equation is asymptotically stable if the
force feedback gain is sufficiently small--- see Appendix
\ref{apx_ForceFB}.

%-------------------------------------------------------------
\section{Calculating States of the Flexible Robot}
\label{sec_flexible}
%-------------------------------------------------------------

To implement control law (\ref{eq_tau'}) the values of $q_r$,
$q_s$, and their time derivatives are required. Vectors $q_r$ and
$\dot q_r$ can be measured, but $q_s$ and its time derivative
should be obtained by calculation. Consider the rheonomic
constraint ${\Phi}(q_r(t),q_s) =0$ whose differentiation with
respect to time leads to
\begin{equation} \label{eq_VelConst}
J_s \dot q_s= J_r \dot q_r (t) = \dot p_r(t)
\end{equation}
In the case that both robots have the same number of degrees of
freedom, i.e., $m =n$, one can obtain $\dot q_s$ uniquely from the
above equation. Subsequently, $q_s$ can be found algebraically
through solving inverse kinematics $\Phi(q_r(t),q_s)=0$, that is
$q_s = \Omega(q_r(t))$. Although, more often an explicit function
may not exist, one can solve the set of nonlinear constraint
equations numerically, e.g., by using the Newton-Raphson method,
as elaborated in Appendix \ref{apx_Newton-Raphson}. Alternatively,
an estimation of $q_s$, can be solved by resorting to a
closed-loop inverse kinematic (CLIK) scheme based on either the
Jacobian transpose \cite{Wolovich-Elliott-1984,Chiacchio-1991} or
the Jacobian {\em pseudoinverse} \cite{Balestrino-Maria-1984}.

%=============================================================
\begin{figure}
\centering{\includegraphics[width=9cm]{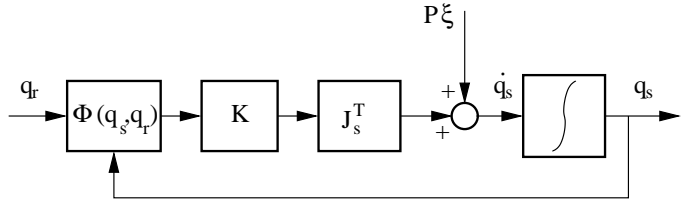}}
\caption{The CLIK jacobian transpose scheme for extracting the
generalized coordinate of the target manipulator.}
\label{fig_CLIK}
\end{figure}
%=============================================================

Since the emulating robot is structurally rigid while the simulated
robot is usually flexible, we can say $n < m$. Therefore, there are
less equations than unknowns in (\ref{eq_VelConst}). The theory of
linear systems equations establishes \cite{GarciadeJalon-Bayo-1994}
that the general solution can be expressed by

\begin{equation} \label{eq_General}
\dot q_s= \dot q_s^p + P  \xi,
\end{equation}
where
\begin{equation} \label{eq_Particular}
\dot q_s^p := J_s^+ \dot p_r(t)= J_s^+ J_r \dot q_r(t).
\end{equation}
is the particular solution and $P \xi$ represents the general
solution of the homogenous velocity. In the above $J_s^+=J_s^T (J_s
J_s^T)^{-1}$ is the pseudoinverse of $J_s$, and $P$ is a projection
matrix whose columns annihilate the constraints
\cite{GarciadeJalon-Bayo-1994}, i.e., $J_s P =0$, and $\xi \in
\mathbb{R}^{n-m}$ is the independent coordinate, which represents
the self-motion of the robot.

Similarly, the general solution of acceleration can be found from
the second derivative of the constraint equations as
\begin{equation}\label{eq_ddqis}
 \ddot q_s = P \dot \xi + J_s^+ c
\end{equation}
with $c$ defined  previously below (\ref{eq_Lambda}). Now,
substituting the acceleration from (\ref{eq_ddqis}) into
(\ref{eq_SimRobot}) and premultiplying by $P^T$ and knowing that
that $P^T$ annihilates the constraint force, i.e., $P^T J_s^T=0$,
we arrive at
\begin{equation} \label{eq_Nddxi}
N \dot \xi = P^T h_s + v,
\end{equation}
where $N= P^T M_s P$ and $v= P^T M_s J_s^+ c$. Equation
(\ref{eq_Nddxi}) expresses the acceleration of the independent
coordinates in closed-form that constitutes the dynamics of the
independent coordinates.

Now, one can make use of the dynamics equation to obtain the
velocity of the independent coordinate $\xi$ by numerical
integration; that, in turn, can be substituted in (\ref{eq_General})
to yield the velocity $\dot q_s$. Subsequently, $q_s$ can be
obtained as a result of numerical integration. However, since
integration inevitably leads to drift, we need to correct the
generalized coordinate, e.g., by using the Newton-Raphson method, in
order to satisfy the constraint precisely. Alternatively, an
estimate of the generalized coordinates of the target robot can be
solved by resorting to a closed-loop inverse kinematics (CLIK)
scheme, where the inverse kinematics problem is solved by
reformulating it in terms of the convergence of an equivalent
feedback control systems
\cite{Chiacchio-1991,Wolovich-Elliott-1984}. Fig.\ref{fig_CLIK}
illustrates the Jacobian transpose CLIK scheme to extract the
generalized coordinates of a redundant system whose null-space
component $P  \xi$ and $q_r$ are considered as input channels. The
feedback loop after the integration of the velocity ensures that
constraint error can be diminished by increasing the feedback gain
\cite{Chiacchio-1991}. More specifically,

\begin{theorem} \label{th_CLIK}
Assume that a gain matrix $K$ is positive-definite and that the
Jacobian matrix $J_s$ is of full rank. Then, the closed-loop inverse
kinematics solution based on a control law $\dot q_s=P \xi + J_s^T K
\Phi(q_s,q_r(T))$, see Fig.\ref{fig_CLIK}, ensures that the
constraint error is ultimately bounded into an attractive ball
centered at $\Phi=0$; the radius of the ball can be reduced
arbitrarily by increasing the minimum eigenvalue of the gain matrix.
\end{theorem}
The proof of the theorem is given in Appendix \ref{apx_CLIK} and
is similar to the proof of convergence of the CLIK solution for
redundant manipulators presented by Chiacchio {\em et al.}
\cite{Chiacchio-1991}.

To summarize, the robot emulation can proceed in the following
steps:

\begin{enumerate}
\item Start at the time when the entire system states, i.e., $\{
q_{s}, \dot q_s, q_r , \dot q_r \}$, are known.

\item Apply the control law (\ref{eq_tau'}). \label{step1}

\item Upon the measurement of the emulating-robot joint angles and
velocities, simulate independent states of the target robot by
making use of the acceleration model (\ref{eq_Nddxi}).

\item Solve the inverse kinematics, e.g., by using the CLIK, to
solve the generalized coordinate of the target robot; then go to
step \ref{step1}.

\end{enumerate}

\begin{figure}
\centering{\includegraphics[width=9cm]{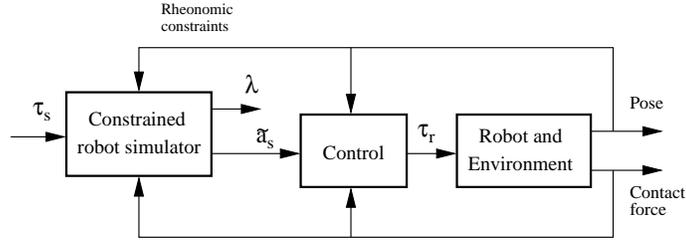}}
\caption{The robot-emulation scheme with the application of the
rheonomic constraint; the simulator comprises the DAE model of the
target robot.} \label{fig_HLS2_system}
\end{figure}

Fig.\ref{fig_HLS2_system} illustrates the implementation of the
robot-emulation scheme with the application of the rheonomic
constraint. Note that the target robot is modelled as a
rheonomically constrained manipulator and the equations of motion
are represented by the DAE system
(\ref{eq_SimRobot})-(\ref{eq_Phi}).

%=============================================================
%=============================================================

\subsection{Flexible-Joint Case Study}
This subsection is devoted to derivation of the emulation scheme
pertaining to robot with {\em flexible-joints}. The generalized
coordinates of the simulated flexible-joint robot are given by
${q}_s^T=[{q}_{s1}^T \;\; {q}_{s2}^T]$, where vectors $q_{s1}$ and
$q_{s2}$ represent the joint angles and motor angles,
respectively. Since only the joint angles contribute to the robot
end-effector motion, we have $\Phi(q_{s1},q_r)=0$, which implies
that the Jacobian of flexible-joint manipulators should be of this
form
\begin{equation} \label{eq_Js1}
J_s= \left[ \begin{array}{cc} J_{s1} & 0 \end{array} \right]
\end{equation}
Assuming submatrix ${J}_{s1}$ is invertible means that the joint
variables can be derived uniquely from the mapping below
\begin{equation}
\dot q_{s1} = J_{s1}^{-1} J_r \dot q_r.
\end{equation}
The independent coordinates, which are comprised of the motor
rotor angles, should be obtained through simulation. The structure
of Jacobian (\ref{eq_Js1}) specifies the projection matrix as
\begin{equation} \label{eq_Proj}
P= \left[ \begin{array}{c}  0 \\ I \end{array} \right].
\end{equation}
Moreover, the dynamics model of flexible-joint manipulators
\cite{Tomei-1991,Jankowski-Elmaraghy-1991} can be written as
\begin{equation} \label{eq_EqMotion}
\left[ \begin{array}{cc} M_{s1} & M_{s12} \\ M_{s12}^T & M_{s2}
\end{array} \right] \left[ \begin{array}{c} \ddot q_{s1} \\ \ddot
q_{s2} \end{array} \right] = \left[ \begin{array}{c} h_{s1} \\
h_{s2} \end{array} \right] + \left[ \begin{array}{c} 0 \\ \tau_s
\end{array} \right]
\end{equation}
Finally, substituting the parameters from (\ref{eq_Proj}) and
(\ref{eq_EqMotion}) into (\ref{eq_Nddxi}) yields $N=M_{s2}$ and
$v=M_{s21}J_{s1}^{-1}c$. Hence, the acceleration model of  the
independent coordinates is given by
\begin{equation} \label{eq_ddqs2}
\ddot q _{s2} = \dot \xi =  M_{s2}^{-1} ( h_{s2} + \tau_s +
M_{s12} J_{s1}^{-1} c ),
\end{equation}
which can be numerically integrated  to obtain $\dot q_{s2}$.

%-------------------------------------------------------------
\subsection{Simulation Without Applying the Constraints}
\label{sec_NoConstraint}
%-------------------------------------------------------------

In this section, we reformulate the robot emulation problem
without imposing the kinematic constraint. Indeed, one can show
that the control law can maintain the asymptotic stability of the
constraint condition if the dynamics model of the emulating robot
is accurately known. Assume that the constraint is no longer
applied to the simulated robot. In other words, the dynamics of
the target robot is described by (\ref{eq_SpaceRobot}), which is
an Ordinary Differential Equation (ODE). Then, we can obtain the
complete states of the target robot from integration of the
acceleration
\[ \ddot q_s= M_s^{-1} ( B \tau_s - h_s + J_s^T f_a). \]
It can be readily shown that the dynamics of the constraint error
under the control law (\ref{eq_tau'}) is described by
\begin{equation} \label{eq_PolyPhi}
\ddot \Phi + G_v \dot \Phi + G_p \Phi =0.
\end{equation}
Therefore, by selecting adequate gains, the constraint error is
exponentially stable, $\Phi, \dot \Phi \rightarrow 0$ as $t
\rightarrow \infty$.

\begin{figure}
\centering{\includegraphics[width=9cm]{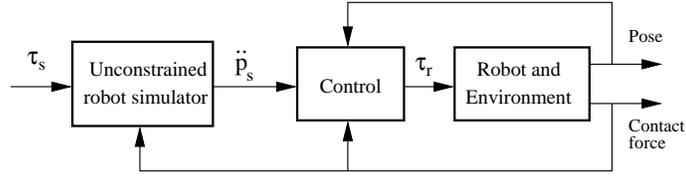}}
\caption{The robot-emulation scheme without the application of the
constraint; the simulator comprises the ODE model of the target
robot.} \label{fig_HLS1_system}
\end{figure}

Fig.\ref{fig_HLS1_system} illustrates the concept of robot emulation
without the application of the constraint. Note that the simulator
contains the dynamics model of the target robot in the ODE form.

%=============================================================
%=============================================================

%-------------------------------------------------------------
\subsection{Topological Difference Between the Two Robot-Emulation Schemes}
\label{sec_Topology}
%-------------------------------------------------------------

Both emulation schemes establish identical force/motion dynamics
between the end-effector of the emulating-robot and that of the
target robot. The difference between the two schemes, however, lies
in how the simulator is synchronized with the actual emulating
robot. In the first scheme, Fig.\ref{fig_HLS2_system}, the
end-effector pose of the simulated robot is rheonomically
constrained with that of the actual robot, and the controller
matches trajectories of the simulated constraint force, i.e. the
Lagrange multiplier, and that of physical constraint force. On the
other hand, in the second scheme, Fig.\ref{fig_HLS1_system}, the
physical constraint force is input to the simulator and the
controller matches the pose trajectories of the simulated and the
actual robots.  The former scheme requires that the simulated robot
be rheonomically constrained and hence the simulator comprises a DAE
system, whereas the latter scheme does not require the application
of the constraint and hence the simulator comprises an ODE system.
Ideally, both schemes should yield the same result. But in presence
of lumped uncertainty and external disturbance, as shall be
discussed in the next section, the first and the second schemes
result in pose error and constraint force error, respectively.

%-------------------------------------------------------------
\section{Stability and Performance Analysis}
\label{sec_Analysis}
%-------------------------------------------------------------

In this section, we limit our attention to stability and
performance analysis of robot emulation in presence of external
disturbance or actuator dynamics to the case where the manipulator
is fully constrained by environment.

Since there is small displacement during the contact phase, the
velocities and centripetal and Coriolis forces go to zero and also
the mass inertia matrices and the Jacobian matrices can be assumed
constant. Hence we will proceed with the {\em frequency domain}
analysis.

%-------------------------------------------------------------
\subsection{Effect of Disturbance}
\label{sec_Disturbance}
%-------------------------------------------------------------

In practice, the fidelity of the robot-emulation is adversely
affected by disturbance and sensor noise. Assume vector $d$ captures
the disturbance caused by error in contact force/moment measurement,
sensor noise, and lumped uncertainty in the robot parameters such as
uncompensated gravitational forces and joint frictions. Then, the
robot dynamics in presence of the disturbance can be written as
\begin{equation} \label{eq_dis}
M_r \ddot{q}_r + h_r = \tau_r + J_r^T (f_a + d).
\end{equation}
In the following, the effect of the additive disturbance on the
two schemes shall be investigated. In the case of the
robot-emulation without application of the constraint, the
dynamics of the constraint error (\ref{eq_PolyPhi}) is no longer
homogeneous. The transmissivity from the disturbance to the
constraint error can be obtained in the Laplace domain by
\begin{equation} \label{eq_ConstError}
\Phi = s^2 Z(s)^{-1} {\cal M}_r^{-1} d,
\end{equation}
where $Z(s)= \frac{s^2}{s^2+ G_v s + G_p}$ can be interpreted as
the admittance of the motion controller.  It is apparent from
(\ref{eq_ConstError}) that decreasing the inertia of the
emulating-robot makes the constraint error more sensitive to the
disturbance and the sensor noise.

In contrast, the robot emulation scheme with the application of the
constraint naturally fulfills the constraint condition even in the
presence of external disturbance. However, the simulated and
emulating robots may no longer experience identical constraint
forces. One can obtain the transmissivity from the disturbance to
the force error in the Laplace domain by
\begin{equation} \label{eq_ForceError}
e_f = - Z(s) {\cal Q}  d,
\end{equation}
where  $ {\cal Q} = {\cal M}_s {\cal M}_r^{-1}$ is the inertia ratio
of the two robots. Adequate choice of gains implies that $\| Z
\|_{\infty}=1$. Hence, we can say $\| e_f \| \leq \bar \sigma ({\cal
Q}) \| d \|$, where $\bar \sigma(\cdot)$ denotes the maximum {\em
singular value}.

\begin{remark} \label{rm_disturbance}
The magnitude of the sensitivity of the force error to the
disturbance depends on the inertia ratio; the higher the ratio $\|
{\cal Q} \|$, the more sensitivity. Roughly speaking, choosing a
massive emulating robot is preferable because it minimizes the
disturbance sensitivity of the emulating system to sensor noise and
disturbance.
\end{remark}

%-------------------------------------------------------------
\subsection{Actuator Dynamics and Contact Stability}
\label{sec_Stability}
%-------------------------------------------------------------

Practical implementation of the control law requires ideal
actuators. In free space, the actuator dynamics can be neglected,
because the fast dynamics of the actuators is well masked by
relatively slow dynamics of the manipulator. In contact phase,
however, where the velocity tends to be zero, the manipulator loses
its link dynamics. In this case, the simulator and actuators form a
closed-loop configuration which may lead to instability.

In the contact phase, the emulating-robot develops constraint
force/moment as a result of its joint torques, i.e.  $f_a = -
J_r^{-T} \tau_r$. The joint torques, in turn, are a function of the
contact force according to the torque-control law (\ref{eq_tau'}).
This input-output connection forms an algebraic loop which cannot be
realized in a physical system unless the actuator dynamics is taken
into account. In fact, the emulating-robot cannot produce joint
torque instantaneously due to finite bandwidth of its actuators and
time delay. Let transfer function $H(s)$ represent the actuator
dynamics, i.e., $\tau_r=H(s) \tau_r^c$, where $\tau_r^c$ is the
joint torque command. Then, the contact force developed by the
emulating-robot is
\begin{equation} \label{eq_Hu}
f_a = - H(s) J_r^{-T} \tau_r^c
\end{equation}
Assuming negligible velocity and acceleration at the contact, i.e.
$\dot q_r \approx 0$, and substituting the torque-control law
(\ref{eq_tau'}) into (\ref{eq_Hu}) yields
\begin{eqnarray} \nonumber
f_a &=& H(s)f_a  -  H(s) Z(s)^{-1} {\cal Q}^{-1} f_a - H(s)
Z(s)^{-1} {\cal Q}^{-1} \lambda \\ \label{eq_FrH} & =& -L(s) ( f_a
- \lambda ),
\end{eqnarray}
where $L(s)= {\cal Q}^{-1} L'(s)$ is the loop-gain transfer
function and $L'(s)= H(s)/(Z(s) - Z(s)H(s))$; recall that $Z(s)$
and ${\cal Q}$ have been already defined in the previous section.
It is clear from (\ref{eq_FrH}) that the actuator dynamics is
responsible to form a force feedback loop which only exist if the
actuator has finite bandwidth, otherwise the force terms from both
sides of equation (\ref{eq_FrH}) are cancelled out.

According to the Nyquist stability criterion for a multi-variable
system \cite{Skogestad-1996}, the closed-loop system is stable if
and only if the Nyquist plot of $\det(I +  L(s))$ does not encircle
the origin. This is equivalent to saying that there must exist a
gain $\epsilon \in (0 \; 1]$ such that \cite{Skogestad-1996}
\begin{equation} \label{eq_det}
\det \left( I + \epsilon L(s) \right) \neq 0 \;\;\;\;\; \forall
s>0
\end{equation}
Since $\det(I + \epsilon L )= \prod_i \lambda_i(I + \epsilon L )$,
$\lambda_i(I + \epsilon {\cal Q}^{-1}  L' )= 1 + \epsilon L'
\lambda_i({\cal Q}^{-1})$, and $\lambda_i({\cal
Q}^{-1})=1/\lambda_i({\cal Q})$, the stability condition
(\ref{eq_det}) is equivalent to the following
\begin{equation} \label{eq_L+1}
\epsilon L'(s) / \lambda_i({\cal Q}) + 1 \neq 0 \;\;\;\;\; \forall
s>0, \; \forall i, \; \forall \epsilon \in (0, 1]
\end{equation}
Let the actuator dynamics (or time delay) be captured by a
first-order system $H(s) = \frac{1}{1 + \omega_a^{-1} s}$, where
$\omega_a$ is the actuator's bandwidth, i.e.,
\[ L'(s)=\frac{s^2+G_v s +G_p}{\omega_a^{-1} s^3}.\]
Then, the characteristic equation of (\ref{eq_L+1}) is given by
\[ \frac{\lambda_i ({\cal Q})}{\epsilon   \omega_a } s^3 + s^2 + G_v s + G_p \neq 0
\;\;\;\;\; \forall s>0, \; \forall i, \; \forall \epsilon \in (0, 1]
\] Since the inertia matrices are always positive definite, one
expects that matrix ${\cal Q}$ to be positive definite too. Assume
that $\lambda_i({\cal Q})>0$, $G_p>0$, and $G_v >0$. Then, according
to the {\em Routh-Hurwitz} stability criterion \cite{Evans-1954},
one can show that the above equation has no roots in the right-half
plane---the schedule of the characteristic equation is written in
Appendix \ref{apx_Routh}---if and only if $\lambda_i({\cal Q})>0$
and
\begin{eqnarray} \nonumber
& & - \lambda_i({\cal Q}) ( \epsilon \omega_a )^{-1} G_p
+ G_v >0  \;\;\;\;\; \forall i, \; \forall \epsilon \in (0, 1] \\
\label{eq_stability} & \Leftrightarrow & \lambda_{max}({\cal Q}) <
\frac {2 \omega_a} {\omega_p},
\end{eqnarray}
where $\omega_P = 2 G_p/ G_v$ is the bandwidth of the controller.
The above development can be summarized in the following
proposition
\begin{proposition}
Assume matrices ${\cal M}_r$ and ${\cal M}_s$ represent the
Cartesian inertia of the emulating and the target robots,
respectively. Also assume that the control law (\ref{eq_tau'})
with bandwidth $\omega_p$ be applied to the emulating-robot whose
actuators have bandwidth $\omega_a$. Then, the emulating-robot can
establish a stable contact if matrix ${\cal Q}={\cal M}_s {\cal
M}_r^{-1}$ remains positive-definite and its maximum eigenvalue is
bounded by $2\omega_a/\omega_p$.
\end{proposition}

\begin{remark} \label{rm_stability}
The stability condition (\ref{eq_stability}) imposes a lower bound
limit on the ratio of Cartesian inertia matrix of the emulating
robot to that of the simulated robot.
\end{remark}

\begin{remark} \label{rm_conflict}
Since $| \lambda_{max}({\cal Q})| \leq \bar \sigma({\cal Q})$,
remarks \ref{rm_disturbance} and \ref{rm_stability} imply that the
low disturbance sensitivity and contact stability requirements
relating to ${\cal Q}$ are not in conflict.
\end{remark}

\subsection{Force Transmissivity}

%=============================================================
\begin{figure}
\centering{\includegraphics[width=8.5cm]{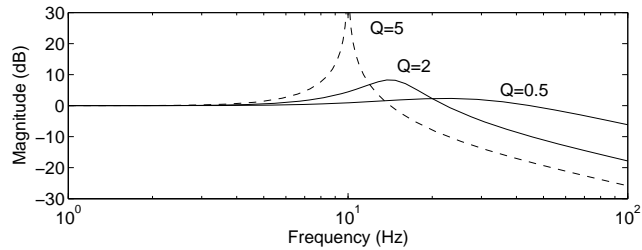}}
\caption{Graph of $|T(j\omega)|$.} \label{fig_TS}
\end{figure}
%=============================================================
The Lagrange multiplier is no longer consistent with ${\lambda}
\equiv f_a$ in presence of actuator dynamics. Assuming stable
contact, one can show from (\ref{eq_FrH}) that the constraint
force $f_a$ is related to the Lagrange multiplier $\lambda$ by
\[ f_a = T(s) \lambda \]
where
\begin{eqnarray} \nonumber
T(s) &=& [I + L(s)]^{-1} L(s) \\ \label{eq_T(s)}  &=& [{\cal Q} +
L'(s)I ]^{-1} L'(s)
\end{eqnarray}
is the transmissivity transfer function.
\begin{remark}
Observe that in the presence of actuator dynamics, the force
response of the emulating-robot depends on the inertia ratio of the
two robots.
\end{remark}

A high fidelity emulation requires that $|T(s)|$ be flat over a wide
bandwidth and then roll-off at high frequency. The graph of $|T(j
\omega)|$ for a scaler case is shown in Fig.\ref{fig_TS} for
$\omega_p=10$, $\omega_a=25$, and different values of ${\cal Q}$.
Note that according to (\ref{eq_stability}), contact stability
requires ${\cal Q} < 5$. It is evident from the figure that ${\cal
Q}$ has to be sufficiently larger than the critical value ${\cal Q}
= 5$ in order to avoid the resonance.

%-------------------------------------------------------------
\section{Conclusion}
\label{sec_conclusion}
%-------------------------------------------------------------

We have presented the development of a methodology for emulation of
a target robot, say a space robot, operating in a complex
environment by using an actual robot. Although the actual
emulating-robot is not dynamically equivalent to the target robot,
the former is controlled such that the dynamic similarity is
achieved. The emulation scheme aims to replicate the dynamic
behavior of the target robot interacting with environment, without
dealing with a complex calculation of the contact dynamics. In our
approach, the target robot is modelled as a rheonomically
constrained manipulator that constitutes a real-time simulator,
which together with the control system, drives the actual robot. It
has been shown that the Lagrange multipliers corresponding to the
constraint is tantamount to a constraint force seen by the simulated
robot. Consequently, the difference between the Lagrange multipliers
and the physical constraint force is chosen as the force error to be
minimized by the controller. To this end, an emulating-robot
controller has been developed such that it (i) produces the endpoint
motion of the target robot and (ii) regulates the force error to
zero.

The fidelity and the stability of the emulation scheme has been
analyzed. In this regard, the inertia ratio of the two robots,
${\cal Q}={\cal M}_s {\cal M}_r^{-1}$, turned out to be an important
factor. Specifically, the analysis has yielded the following
results:
\begin{enumerate}
\item{ $\|{\cal Q} \| \gg 1 $ makes the emulating system sensitive
to force/moment sensor noise and disturbance; a low-noise
force/moment sensor is required.}

\item{ $\lambda_{max}({\cal Q}) \gg 1$ may lead to contact
instability due to actuator dynamics and delay; a high bandwidth
actuator is required.}

\item{ $\|{\cal Q} \| = 1$ minimizes actuation effort; a large
deviation from one requires actuators with large torque capacity.}
\end{enumerate}
Theses remarks should be considered in design of the emulating
robot.

In addition, a number of preliminary experiments for emulation of
flexible-joint robots have been performed to support the concept
of robot-emulation.

\appendix
%-------------------------------------------------------------
\section{}
\label{apx_ForceFB}
%-------------------------------------------------------------
Choose the following Lyapunov function
\[ V_1 = \frac{1}{2} \dot e_p^T  \dot e_p +  \frac{1}{2} \dot e_p^T  G_f {\cal M}_s \dot e_p + \frac{1}{2}
e_p^T G_p e_p, \] whose time derivative is
\[\dot V_1 = - \frac{1}{2} \dot e_p^T (2 G_v I - G_f \dot{\cal
M}_s) \dot e_p. \] Since $\lambda_i(2 G_v I - G_f \dot{\cal M}_s) =
2 G_v - G_f \lambda_i(\dot{\cal M}_s)$, we can say $\dot V_1 < 0$ if
\[ G_f < \frac{2 G_v }{\lambda_{max}(\dot{\cal M}_s)}. \]

%-------------------------------------------------------------
\section{}
\label{apx_Newton-Raphson}
%-------------------------------------------------------------

The constraint equations  can be  linearized up to a first-order
approximation by writing the Taylor expansion around a
neighborhood, say initial guess $q_s^0$, as
\begin{equation} \label{eq_approx}
\Phi(q_s,q_r) + \Phi_{q_s} (q_s - q_s^0 ) \approx 0.
\end{equation}
Where the variables $q_r$ is considered as a constant value, and
$\Phi_{q_s} = -J_s$. Then, the following loop
\[ q_s^{k+1} = q_s^k + J_s^+ {\Phi}
(q_s^k , q_r) \] can be worked out iteratively until the error in
the constraint falls within an acceptable tolerance range, e.g.
$\|\Phi \| \leq \epsilon$.

%-------------------------------------------------------------
\section{}
\label{apx_CLIK}
%-------------------------------------------------------------
Consider the positive definite function candidate
\[ V= \frac{1}{2} \Phi^T K \Phi. \]
Its derivative along the trajectories of the system is
\begin{eqnarray} \label{eq_DV}
\dot V &=& \Phi^T K \dot p_r - \Phi^T K J_s J_s^T K \Phi - \Phi^T
K J_s P  \xi \\ \nonumber & =&  \Phi^T K \dot p_r - \Phi^T K J_s
J_s^T K \Phi.
\end{eqnarray}
where the last term in the right-hand side of (\ref{eq_DV})
vanishes because $J_s P =0$. An upper bound of $\dot V$ is
\begin{equation} \label{eq_V}
\dot V \leq \| \Phi \| \lambda_M(K) \| \dot p_r \| - \| \Phi \|^2
\lambda_m^2(K) \lambda_m (J_s J_s^T).
\end{equation}
Denoting $\lambda_M(A)$ and $\lambda_m(A)$ as maximum and minimum
eigenvalues of matrix $A$, we can find the following condition for
negative $\dot V$
\begin{equation} \label{eq_dV} \dot V \leq 0 ,
\;\;\;\;\; \forall \| \Phi \| \geq \mu= \frac{ \lambda_M(K) \|\dot
p_r \|_{max}}{\lambda^2_m(K) \lambda_m(J_s J_s^T)}.
\end{equation}
Furthermore, we can say the following bounds on the Lyapunov
function as
\[ \frac{1}{2} \lambda_m(K) \| \Phi \|^2 \leq V \leq \frac{1}{2} \lambda_M(K) \| \Phi \|^2\]
Therefore, according to the uniform ultimate boundedness theorem of
uncertain dynamic system \cite{Corless-Letmann-1981,Khalil-1992},
one can say from (\ref{eq_V}) and (\ref{eq_DV}) that the constraint
error $\Phi$ is then uniformly ultimately bounded into the ball
centered at $\Phi=0$ and radius
\[ \rho = \frac{ \gamma^{3/2}(K) \| \dot p_r \|_{max}}{
\lambda_m(K) \lambda_m(J_s J_s^T)}, \] where $\gamma(A) =
\lambda_M(A) / \lambda_m(A)$ is the condition number of matrix A,
which concludes the proof. Note that $K$ can be conveniently chosen
as a diagonal matrix $K=k I$, and hence $\gamma = 1$ and
$\lambda_m(K)=k$. Then, the error can be diminished at will by
increasing $k$ \cite{Canudas-Siciliano-Bastin-book-1996}.

%-------------------------------------------------------------
\section{}
%\section{The Routh-Hurwitz Schedule}
\label{apx_Routh}
%-------------------------------------------------------------
The {\em Routh-Hurwitz} schedule of the coefficients of the
characteristic equation is
\[
\begin{array}{rl}
{\begin{array}{l} s^3 \\ s^2 \\ s \\ 1 \end{array}} &
{\left|\begin{array}{ll}
\lambda_i({\cal Q})/\epsilon \omega_a & G_v \\
 1 & G_p \\ G_v - \lambda_i ({\cal Q})G_p / \epsilon \omega_a & 0 \\ G_p & 0
\end{array} \right.}
\end{array} \]
The Routh-Hurwitz criterion states that the number of roots of the
characteristic equation with positive real parts is equal to the
number of changes in the sign of the first column array.

%-------------------------------------------------------------
\section{}
%\section{The Flexible-Joint Arm}
\label{apx_FlexibleJoint}
%-------------------------------------------------------------

In this appendix, the control algorithm for emulation of 1-dof
flexible joint arm is developed. The dynamics of a 1-dof flexible
joint arm can be expressed by
\cite{Tomei-1991,Jankowski-Elmaraghy-1991}
\[  \!\!  \left[ \begin{array}{c} m_{s1} \ddot{q}_{s1}  \\  m_{s2}\ddot{q}_{s2} \end{array} \right] \!\!
= \!\! \left[ \begin{array}{c} k(q_{s2} - q_{s1}) + \zeta (\dot{q}_{s2} - \dot{q}_{s1})  \\
-k(q_{s2} - q_{s1} ) - \zeta (\dot{q}_{s2} - \dot{q}_{s1} ) -c_s \dot{q}_{s2} + \tau_s(t)
\end{array} \right]  \]
where $m_{s1} = 0.05 kgm^2$ and $m_{s2} = 0.1 kgm^2$ are the link
inertia and the drive inertia, $l= 0.3 m$ is the arm length, $k =
3.0 Nm/rad$ represents joint stiffness, $\zeta = 0.1 Nm.s/rad$ is
damping at joint, and $\tau_s(t)$ is real-time control input of
the simulated robot -- in this particular illustration, the
control law is comprised of the resolved-rate controller
accompanied with a force-moment accommodation (FMA) loop
\cite{Aghili-Dupuis-Martin-Piedboeuf-2001a}.

The dynamics of the rigid joint robot is  simply described by
\[ m_r \ddot{q}_r + c_r \dot{q}_r + w \cos(q_r) = \tau_r + fl \]
where $c_r =0.022 Nm.s/rad$ represents joint viscous friction, and
$m_r =0.05 kgm^2$ is the collective arm inertia, and $w=0.5 Nm$ is
the maximum gravity torque. Since $J_{s1}=J_r=l$, we have $q_{s1}
\equiv q_r$. Hence, the control law for the rigid robot makes it
behave as a flexible joint robot whose dynamics is expressed by
\begin{eqnarray} \nonumber \tau_r &=& c_r \dot q_r + w \cos (q_r)
- f l + m_r m_{s1}^{-1} [ \ddot q_{s1} \\ \nonumber &+& G_v( \dot
q_r - \dot q_{s1}) + G_p (q_r - q_{s1})],
\end{eqnarray}
where
\[ \dot \xi = m_{s2}^{-1}[-k( \xi - q_r ) - \zeta
(\dot \xi - \dot{q}_r) + \tau_s(t)], \] and
\[ \ddot q_{s1}= m_{s1}^{-1}[k(\xi - q_r) + \zeta(\dot \xi - \dot
q_r)]. \]

%-------------------------------------------------------------
\bibliographystyle{IEEEtran}
%\bibliography{references}

\begin{thebibliography}{10}

\bibitem{Aghili-Piedboeuf-2006}
F.~Aghili and J.-C. Piedboeuf, ``Emulation of robots interacting with
  environment,'' \emph{IEEE/ASME Trans. on Mechatronics}, vol.~11, no.~1, pp.
  35--46, Feb. 2006.

\bibitem{Skaar-Ruoff-Seebass-1994}
S.~B. Skaar, C.~F. Rouff, and A.~R. Seebass, \emph{Teleoperation and Robotics
  in Space}.\hskip 1em plus 0.5em minus 0.4em\relax Washington, DC: American
  Institute of Aeronautics and Astronautics, Inc., 1994.

\bibitem{Aghili-2010n}
F.~Aghili, ``Cartesian control of space manipulators for on-orbit servicing,''
  in \emph{{AIAA} Guidance, Navigation and Control Conference}, Toronto,
  Canada, August 2010.

\bibitem{Aghili-Su-2012}
F.~Aghili and C.-Y. Su, ``Reconfigurable space manipulators for in-orbit
  servicing and space exploration,'' in \emph{International Symposium on
  Artificial Intelligence, Robotics and Automation in Space {i-SAIRAS}}, Turin,
  Italy, Sep.~4--6 2012.

\bibitem{Aghili-2012b}
F.~Aghili, ``Active orbital debris removal using space robotics,'' in
  \emph{International Symposium on Artificial Intelligence, Robotics and
  Automation in Space {i-SAIRAS}}, Turin, Italy, Sep.~4--6 2012.

\bibitem{Piedboeuf-Aghili-Doyon-Martin-2002}
J.-C. Piedb{\oe}uf, F.~Aghili, M.~Doyon, and E.~Martin, ``Dynamic emulation of
  space robot in one-g environment using hardware-in-loop simulation,'' in
  \emph{{CISM-IFToMM} Symposium on Robotics Design, Dynamics and Control},
  Italy, July~3--6 2002.

\bibitem{Aghili-Piedboeuf-2002}
F.~Aghili and J.-C. {Piedb{\oe}uf}, ``Contact dynamics emulation for
  hardware-in-loop simulation of robots interacting with environment,'' in
  \emph{{IEEE} International Conference on Robotics \& Automation}, Washington,
  USA, May 11--15 2002, pp. 534--529.

\bibitem{Aghili-2011k}
F.~{Aghili}, ``A prediction and motion-planning scheme for visually guided
  robotic capturing of free-floating tumbling objects with uncertain
  dynamics,'' \emph{IEEE Transactions on Robotics}, vol.~28, no.~3, pp.
  634--649, June 2012.

\bibitem{Aghili-Salerno-2016}
F.~Aghili and A.~Salerno, \emph{Multisensor Attitude Estimation and
  Applications}, 1st~ed.\hskip 1em plus 0.5em minus 0.4em\relax CRC Press,
  2016, ch. Adaptive Data Fusion of Multiple Sensors for Vehicle Pose
  Estimation.

\bibitem{Aghili-Parsa-2007b}
F.~Aghili and K.~Parsa, ``Adaptive motion estimation of a tumbling satellite
  using laser-vision data with unknown noise characteristics,'' in \emph{2007
  IEEE/RSJ International Conference on Intelligent Robots and Systems}, Oct
  2007, pp. 839--846.

\bibitem{Piedboeuf-Dupuis-deCarufel-1998}
J.-C. {Piedb{\oe}uf}, E.~Dupuis, and J.~{de Carufel}, ``{STVF} concept
  document,'' Canadian Space Agency, St-Hubert, Quebec, Tech. Rep.
  CSA-SS-ST-016, Feb. 1998.

\bibitem{Aghili-Dupuis-Piedboeuf-deCarufel-1999}
F.~Aghili, E.~Dupuis, J.-C. {Piedb{\oe}uf}, and J.~{de Carufel},
  ``Hardware-in-the-loop simulations of robots performing contact tasks,'' in
  \emph{International Symposium on Artificial Intelligence and Robotics \&
  Automation in Space: {i-SAIRAS}}, M.~Perry, Ed.\hskip 1em plus 0.5em minus
  0.4em\relax Noordwijk, The Netherland: ESA Publication Division, 1999, pp.
  583--588.

\bibitem{Aghili-2019c}
F.~{Aghili}, ``Robust impedance-matching of manipulators interacting with
  uncertain environments: Application to task verification of the space
  station's dexterous manipulator,'' \emph{IEEE/ASME Transactions on
  Mechatronics}, vol.~24, no.~4, pp. 1565--1576, Aug 2019.

\bibitem{Piedboeuf-Aghili-Doyon-Gonthier-Martin-Zhu-2001}
J.-C. {Piedb{\oe}uf}, F.~Aghili, M.~Doyon, Y.~Gonthier, E.~Martin, and W.-H.
  Zhu, ``Emulation of space robot through hardware-in-the-loop simulation,'' in
  \emph{The 6th International Symposium on Artificial Intelligence and Robotics
  \& Automation in Space: {i-SAIRAS}}, Canadian Space Agency, St-Hubert,
  Quebec, Canada, Jun.~18--22 2001.

\bibitem{Piedboeuf-deCarufel-Aghili-Dupuis-1999}
J.-C. {Piedb{\oe}uf}, J.~{de Carufel}, F.~Aghili, and E.~Dupuis, ``Task
  verification facility for the {C}anadian special purpose dextrous
  manipulator,'' in \emph{{IEEE} Int. Conf. on Robotics \& Automation},
  Detroit, Michigan, May~10--15 1999, pp. 1077--1083.

\bibitem{deCarufel-Martin-Piedboeuf-2000}
J.~de~Carufel, E.~Martin, and J.-C. {Piedb{\oe}uf}, ``Control strategies for
  hardware-in-the-loop simulation of flexible space robots,'' \emph{IEEE
  Proceedings-D: Control Theory and Applications}, vol. 147, no.~6, pp.
  569--579, 2000.

\bibitem{Ma-Buhariwala-Roger-MacLean-Carr-1997}
O.~Ma, K.~Buhariwala, N.~Roger, J.~MacLean, and R.~Carr, ``{MDSF}- {A} generic
  development and simulation facility for flexible, complex robotic systems,''
  vol.~15, pp. 49--62, 1997.

\bibitem{Vafa-Dubowsky-1990}
Z.~Vafa and S.~Dubowsky, ``The kinematics and dynamics of space manipulators:
  The virtual manipulator approach,'' vol.~9, no.~4, pp. 3--21, Aug. 1990.

\bibitem{Vafa-Dubowsky-1990b}
------, ``On the dynamics of space manipulators using the virtual manipulator,
  with applications to path planning,'' vol.~38, no.~4, pp. 441--472,
  Oct.--Dec. 1990.

\bibitem{Ma-Nahon-1992}
O.~Ma and M.~Nahon, ``A general method for computing the distance between two
  moving objects using optimization techniques,'' \emph{ASME Advances in Design
  Automation}, vol.~1, pp. 109--117, 1992, dE-Vol.44-1.

\bibitem{Lankarani-Nikravesh-1988}
H.~M. Lankarani and P.~E. Nikravesh, ``Application of the canonical equations
  of motion in problems of constrained multibody systems with intermittent
  motion,'' in \emph{{ASME} Design Automation Conference}.\hskip 1em plus 0.5em
  minus 0.4em\relax Orland (Florida): ASME, 1988, aSME-Paper No. 88-DAC-51.

\bibitem{Johnson-1985}
K.~L. Johnson, \emph{Contact Mechanics}.\hskip 1em plus 0.5em minus 0.4em\relax
  London: Cambridge University Press, 1985.

\bibitem{Dubowsky-Durfee-Kuklinski-1994}
S.~Dubowsky, W.~Durfee, A.~Kulinski, U.~M{\"u}ller, I.~Paul, and J.~Pennington,
  ``The design and implementation of a laboratory test bed for space robotics:
  The ves mod {II},'' in \emph{{ASME} Conf. {DE}-Vol. 72, {R}obotics:
  {K}inematics, {D}ynamics and {C}ontrol}, 1994, pp. 99--108.

\bibitem{Krenn-Schafer-1999}
R.~Krenn and B.~Sch{\"a}fer, ``Limitations of hardware-in-the-loop simulations
  of space robotics dynamics using industrial robots,'' M.~Perry, Ed.\hskip 1em
  plus 0.5em minus 0.4em\relax Noordwijk, The Netherland: ESA Publication
  Division, 1999, pp. 681--686.

\bibitem{Ananthakrishnan-Teders-Alder-1996}
S.~Ananthakrishnan, R.~Teders, and K.~Alder, ``Role of estimation in real-time
  contact dynamics enhancement of space station engineering facility,''
  \emph{IEEE Robotics \& Automation Magazine}, no.~3, pp. 20--28, Sep. 1996.

\bibitem{Aghili-2006b}
F.~Aghili, ``A mechatronics testbed for manipulator joints,'' in \emph{{IEEE}
  Int. Conference on Robotics \& Automation}, Orlando, Florida, May 2006, pp.
  2188--2194.

\bibitem{Tarao-Inohira-Uchiyama-2000}
S.~Tarao, E.~Inohira, and M.~Uchiyama, ``Motion simulation using a high-speed
  parallel link mechanism,'' in \emph{The 2000 {IEEE/RSJ} Int. Conf. On
  Intelligent Robots and Systems}, Takamatsu, Japan, 2000.

\bibitem{Aghili-Namvar-2008}
F.~Aghili and M.~Namvar, ``Scaling inertia properties of a manipulator payload
  for 0-g emulation of spacecraft,'' \emph{The International Journal of
  Robotics Research}, vol.~28, no.~7, pp. 883--894, July 2009.

\bibitem{Aghili-2005b}
F.~Aghili, ``A robotic testbed for zero-g emulation of spacecraft,'' in
  \emph{{IEEE/RSJ} Int. Conference on Intelligent Robots and Systems},
  Edmonton, Alberta, Canada, 2005, pp. 1033--1040.

\bibitem{Aghili-Namvar-Vukovich-2006}
F.~Aghili, M.~Namvar, and G.~Vukovich, ``Satellite simulator with a hydraulic
  manipulator,'' in \emph{{IEEE} Int. Conference on Robotics \& Automation},
  Orlando, Florida, May 2006, pp. 3886--3892.

\bibitem{Aghili-Parsa-2009}
F.~Aghili and K.~Parsa, ``Motion and parameter estimation of space objects
  using laser-vision data,'' \emph{{AIAA} Journal of Guidance, Control, and
  Dynamics}, vol.~32, no.~2, pp. 538--550, March 2009.

\bibitem{Aghili-2006}
F.~Aghili, ``A mechatronic testbed for revolute-joint prototypes of a
  manipulator,'' \emph{IEEE Trans. on Robotics}, vol.~22, no.~6, pp.
  1265--1273, Dec. 2006.

\bibitem{Aghili-Parsa-Martin-2008a}
F.~Aghili, K.~Parsa, and E.~Martin, ``Robotic docking of a free-falling space
  object with occluded visual condition,'' in \emph{9th Int. Symp. on
  Artificial Intelligence, Robotics \& Automation in Space}, Los Angeles, CA,
  Feb.~26 -- 29 2008.

\bibitem{Aghili-2016c}
F.~Aghili and C.~Y. Su, ``Robust relative navigation by integration of icp and
  adaptive kalman filter using laser scanner and imu,'' \emph{IEEE/ASME
  Transactions on Mechatronics}, vol.~21, no.~4, pp. 2015--2026, Aug 2016.

\bibitem{Aghili-2008c}
F.~Aghili, ``Optimal control for robotic capturing and passivation of a
  tumbling satellite with unknown dyanmcis,'' in \emph{{AIAA} Guidance,
  Navigation and Control Conference}, Honolulu, Hawaii, August 2008.

\bibitem{Aghili-2010f}
------, ``Automated rendezvous \& docking {(AR\&D)} without impact using a
  reliable 3d vision system,'' in \emph{{AIAA} Guidance, Navigation and Control
  Conference}, Toronto, Canada, August 2010.

\bibitem{Canudas-Siciliano-Bastin-book-1996}
C.~Canudas~de Wit, B.~Siciliano, and G.~Bastin, Eds., \emph{Theory of Robot
  Control}.\hskip 1em plus 0.5em minus 0.4em\relax London, Great Britain:
  Springer, 1996.

\bibitem{Craig-1989}
J.~J. Craig, \emph{Introduction to Robotics: {M}echanical and Control},
  2nd~ed.\hskip 1em plus 0.5em minus 0.4em\relax Reading, Massachusetts:
  Addison-Wesley Publishing Company, 1989.

\bibitem{deJalon-Bayo-1989}
J.~G. de~Jalon and E.~Bayo, \emph{Kinematic and Dynamic Simulation of Multibody
  Systems}.\hskip 1em plus 0.5em minus 0.4em\relax Springer-Verlag, 1989.

\bibitem{Aghili-2005}
F.~Aghili, ``A unified approach for inverse and direct dynamics of constrained
  multibody systems based on linear projection operator: Applications to
  control and simulation,'' \emph{IEEE Trans. on Robotics}, vol.~21, no.~5, pp.
  834--849, Oct. 2005.

\bibitem{Aghili-Piedboeuf-2003a}
F.~Aghili and J.-C. {Piedb{\oe}uf}, ``Simulation of motion of constrained
  multibody systems based on projection operator,'' \emph{Journal of Multibody
  System Dynamics}, vol.~10, pp. 3--16, 2003.

\bibitem{Aghili-Parsa-2008b}
F.~Aghili and K.~Parsa, ``An adaptive vision system for guidance of a robotic
  manipulator to capture a tumbling satellite with unknown dynamics,'' in
  \emph{{IEEE/RSJ} Int. Conf. on Intelligent Robots and Systems}, Nice, France,
  September 2008, pp. 3064--3071.

\bibitem{Spong-Vidasagar-1989}
M.~W. Spong and M.~Vidyasagar, \emph{Robot Dynamics and Control}.\hskip 1em
  plus 0.5em minus 0.4em\relax New York, NY: Wiley, 1989.

\bibitem{Aghili-2010}
F.~Aghili, ``Robust impedance control of manipulators carrying heavy payload,''
  \emph{ASME Journal of Dynamic Systems, Measurements, and Control}, vol. 132,
  September 2010.

\bibitem{Wolovich-Elliott-1984}
W.~A. Wolovich and H.~Elliott, ``A computational technique for inverse
  kinematics,'' in \emph{{IEEE} Conf. On Decision and Control}, New York, 1984,
  pp. 1359--1363.

\bibitem{Chiacchio-1991}
P.~Chiacchio, S.~Chiaverini, L.~Sciavicco, and B.~Siciliano, ``Closed-loop
  inverse kinematics schemes for constrained redundant manipulators with task
  space augmentation and task priority strategy,'' \emph{The Int. Journal of
  Robotics Research}, vol.~10, no.~4, pp. 410--425, Aug. 1991.

\bibitem{Balestrino-Maria-1984}
A.~Balestrino, G.~D. Maria, and L.~Sciavicco, ``Robust control of robotic
  manipulators,'' \emph{Reprint of the 9th IFAC World Congress}, vol.~6, pp.
  80--85, Jul. 1984.

\bibitem{GarciadeJalon-Bayo-1994}
J.~{Garcia de Jal{\'o}n} and E.~Bayo, \emph{Kinematic and Dynamic Simulation of
  Multibody Systems: The Real-Time Challenge}.\hskip 1em plus 0.5em minus
  0.4em\relax New York: Springer-Verlag, 1994.

\bibitem{Tomei-1991}
P.~Tomei, ``A simple {PD} controller for robots with elastic joints,''
  \emph{IEEE Trans. on Automatic Control}, vol.~36, pp. 1208--1213, 1991.

\bibitem{Jankowski-Elmaraghy-1991}
K.~P. Jankowski and H.~A. ElMaraghy, ``Dynamic control of flexible joint robots
  with constrained end-effector motion,'' in \emph{Prepr. 3rd {IFAC} Symp. On
  Robot Control}, Vienna, 1991, pp. 345--350.

\bibitem{Skogestad-1996}
S.~Skogestad and I.~Postlethwaite, \emph{Multivariable Feedback Control
  Analysis and Design}.\hskip 1em plus 0.5em minus 0.4em\relax West Sussex PO19
  1UD, England: John Wiley \& Sons, 1996.

\bibitem{Evans-1954}
W.~R. Evans, ``Control system synthesis by root locus method,''
  \emph{Transactions of the AIEE}, vol.~69, pp. 1--4, 1954.

\bibitem{Corless-Letmann-1981}
M.~J. Corless and G.~Letmann, ``Continous state feedback guaranteeing uniform
  ultimate boundedness for uncertain dynmamic systems,'' \emph{IEEE Trans. on
  Automatic Control}, vol. AC-26, no.~5, pp. 1139--1144, Oct. 1981.

\bibitem{Khalil-1992}
H.~K. Khalil, \emph{Nonlinear Systems}.\hskip 1em plus 0.5em minus 0.4em\relax
  New-York: Macmillan Publishing Company, 1992.

\bibitem{Aghili-Dupuis-Martin-Piedboeuf-2001a}
F.~Aghili, E.~Dupuis, E.~Martin, and J.-C. {Piedb{\oe}uf}, ``{Force/Moment}
  accommodation control for tele-operated manipulators performing contact tasks
  in stiff environment,'' in \emph{Proceedings of the 2001 {IEEE/RSJ}
  International Conference on Intelligent Robotics and Systems}, Maui, Hawaii,
  USA, Oct.~29--Nov.~03 2001, pp. 2227 -- 2233.


\end{thebibliography}
%\end{document}
%-------------------------------------------------------------

\end{document}